\documentclass{article}
\usepackage{spconf,amsmath,graphicx}

\usepackage{enumitem}
\setlist{nosep, leftmargin=14pt}

\usepackage{mwe} 
\usepackage[dvipsnames]{xcolor}
\usepackage{mathbbol}
\usepackage[normalem]{ulem}

\newcommand{\etal}{\textit{et al.}}

\title{
Unsupervised Anomaly Detection on Implicit Shape representations for Sarcopenia Detection}

\name{\begin{tabular}{c}
Louise Piecuch$^{\star}$, Jeremie Huet (MD)$^{\dagger\diamond}$, Antoine Frouin (PT)$^{\dagger}$, \\
Antoine Nordez$^{\dagger\ddagger}$, Anne-Sophie Boureau (MD)$^{\diamond\ddagger}$, Diana Mateus$^{\star}$
\end{tabular}\thanks{}}

\address{
$^{\star}$ Nantes Universit\'e, \'Ecole Centrale Nantes, CNRS, LS2N, UMR 6004, F-44000 Nantes, France \\
$^{\dagger}$ Nantes Universit\'e, Movement - Interactions - Performance, MIP, IP UR 4334 UFR STAPS,  Nantes - France\\
$^{\diamond}$ Nantes Universit\'e, CHU Nantes, CNRS, INSERM, l’institut du thorax, F-44000 Nantes, France \\
$^{\ddagger}$ Institut Universitaire de France  (IUF),   Paris -  France 
}
\begin{document}
\maketitle

\begin{abstract}

Sarcopenia is an age-related progressive loss of muscle mass and strength that significantly impacts daily life. A commonly studied criterion for characterizing the muscle mass has been the combination of 3D imaging and manual segmentations. In this paper, we instead study the muscles' shape. We rely on an implicit neural representation (INR) to model normal muscle shapes. We then introduce an unsupervised anomaly detection method to identify sarcopenic muscles based on the reconstruction error of the implicit model. Relying on a conditional INR with an auto-decoding strategy, we also learn a latent representation of the muscles that clearly separates normal from abnormal muscles in an unsupervised fashion. Experimental results on a dataset of 103 segmented volumes indicate that our double anomaly detection strategy effectively discriminates sarcopenic and non-sarcopenic muscles.
\end{abstract}

\begingroup
\renewcommand{\thefootnote}{}
\footnotetext{\textcopyright{} 2025 IEEE. Personal use of this material is permitted. Permission from IEEE must be obtained for all other uses, in any current or future media, including reprinting/republishing this material for advertising or promotional purposes, creating new collective works, for resale or redistribution to servers or lists, or reuse of any copyrighted component of this work in other works.}
\endgroup

\section{Introduction}
\label{sec:intro}
Sarcopenia is a degenerative disease characterized by muscle fat infiltrations and loss of muscular mass. It primarily affects older adults leading to decreased strength and potential independence loss~\cite{cruz2019sarcopenia}\cite{chianca2022sarcopenia}. 
Current diagnosis guidelines focus on strength tests and muscular mass measurements ~\cite{chianca2022sarcopenia}.
However, there is no agreement on the criteria, and different recommendations exist in Europe and Asia. \cite{kim2024}. 
A recent European consensus \cite{cruz2019sarcopenia} highlights the importance of early detection and treatment, and calls for a multi-parametric muscle appreciation approach. Interestingly, the shape of the muscle, an apparently obvious parameter, has rarely been studied, despite its expected role in force production.

A critical bottleneck in mass-based criteria is the need to segment individual muscles from volumetric images. Several deep-learning solutions have thus focused on assisting the segmentation of muscles from CT and MR, ultrasound and x-ray data~\cite{ piecuch2023muscle,alchanti:tmi2021:ultrasound-segmentation, sato:miccai2023:x-ray}. They target the accurate estimation of muscle volume, a known but weak sarcopenia biomarker~\cite{kim2024}.
Instead, we aim to characterize patients based on their muscular shape. Muscle shape has been characterized through classical shape descriptors~\cite{ward:academic-radiology2007:shape-descriptors,
hajghanbari:academic-radiology2011:shape-descriptors}, or relying on statistical shape models~\cite{ghouth:sci-rep2022:ssms-muscle,sutherland:sports-science2023:ssm}.
However, such methods require heavy preprocessing.
Instead, we propose an implicit representation of the muscles' shape. Moreover, we concieve an anomaly detection approach based on the implicit representation of a ``normal" population. 

In practice, we rely on the implicit neural representation by Amiranashvili~\etal\cite{AMIRANASHVILI2024}, capable of modelling a shape-prior over a population. The method was initially designed to predict entire volumes from 2D sparse annotations, relying on the prior to accelerate manual segmentations. Here, we adapt the implicit shape-prior to an Anomaly Detection (AD) task. Similar to AD approaches based on autoencoders or generative models~\cite{pang:acm2021:review-deep-anomaly-detection}, we train the model with a dataset of non-sarcopenic muscles. Since the learned shape-prior captures the variations only within the training population, the reconstruction error is expected to be higher for out-of-distribution muscles,
enabling the detection of sarcopenic muscles.
Experimental results on a dataset of $103$ segmented volumes detailed in  Sec.~\ref{ssec:expe_settings}, shows that the Dice reconstruction error is effectively lower for sarcopenic muscles. Moreover, a Linear Discriminant Analysis (LDA) of the latent shape representations, naturally clusters the non-sarcopenic training populations,  and clearly separates sarcopenia muscles during inference. This separation, combined with a simple classifier, could assist physicians during  diagnosis. 

\section{Related work}
\label{sec:related_work}
Implicit Neural Representations (INRs) have been shown effective in representing 3D shapes, by letting simple coordinate-based neural networks continuously represent an occupancy grid while fitting a Signed Distance Function (SDF)~\cite{Gropp}. Neural SDFs address known problems of voxel and mesh representations, including limited resolution or foldings. They have been adapted to fit single or multiple surfaces from medical images~\cite{alblas:heart-workshop2022:atlas,AMIRANASHVILI2024}. A clever way to learn a consistent INR across multiple shapes is the autodecoder (AutoD) principle introduced in~\cite{park:cvpr2019:deepsdf} and used in \cite{AMIRANASHVILI2024}. An AutoD conditions the INR to an additional learnable latent variable defined on a coordinate system shared over the dataset. The joint INR  becomes a conditional generative shape model. 
When trained on a population, the AutoD INR learns an implicit shape prior, with latent variables encoding variations across the population, including  changes in shape. Such shape prior was used to complete partial annotations for volume segmentation \cite{AMIRANASHVILI2024}. 
We extend this approach to the unsupervised detection of abnormal shapes, relying either on the reconstruction error or distances in the learned latent space. To the best of our knowledge, anomaly detection based on INRs has been studied for images~\cite{naval:miccai2021:anomaly-images} and to detect abnormal regions in point-clouds~\cite{bergmann:wacv2023:anomaly-points}, but not for global shape abnormalities.

\section{Method}
\label{sec:methodo}
\begin{figure}[bt!]
  \centering
  \includegraphics[width=8.5cm]{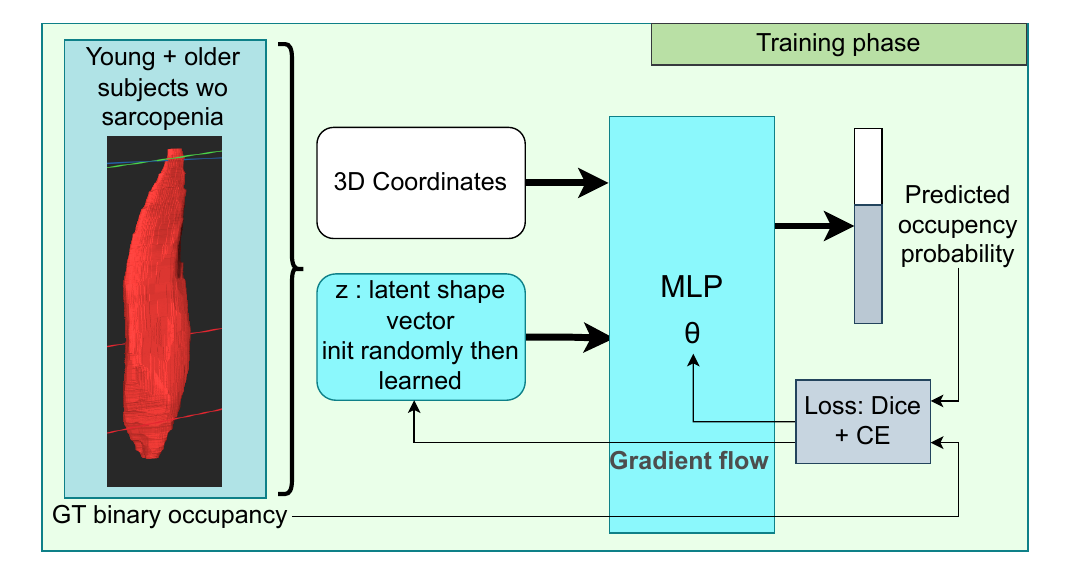} 
  \includegraphics[width=8.5cm]{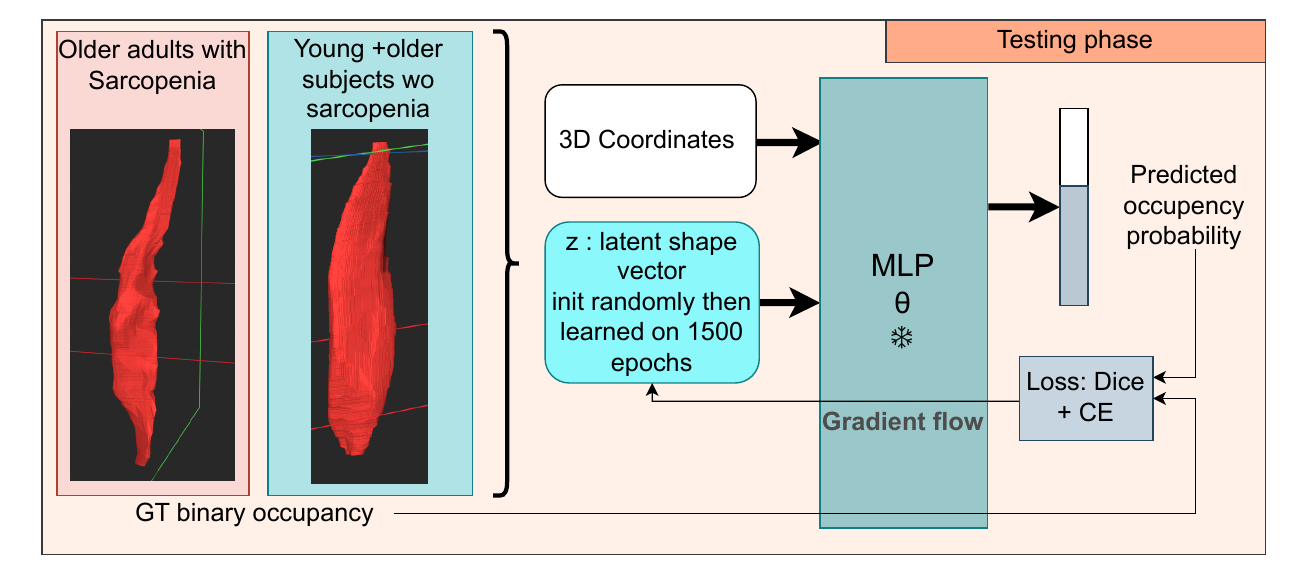} 
  \caption{Overview of the training and testing phases.}
  \label{fig:train-and-test-phases}
\end{figure}
This paper aims to develop a generative model to distinguish variations between ``normal'' and sarcopenic muscles. Given a dataset of $N$ segmented ``normal" muscles $\{Y_i\}_{i=1}^{N}$, we train a generative model $f_\theta$ to learn the expected normal shape variations across the population. Here, each muscle $Y_i$ is a binary volumetric occupancy grid.
We specifically choose the implicit generative model in \cite{AMIRANASHVILI2024} since it provides for each input volume: a latent representation $z_i$ in a shared coordinate system, and a reconstruction of the original input volume $\hat Y_i$. As later shown, applying $f_\theta$ to reconstruct abnormal (i.e. sarcopenic) data, leads to observable discrepancies both in $z_i$ and $\hat Y_i$, with respect to the ``normal'' training data. Next, we describe in detail the chosen generative model, as well as the training and inference phases illustrated in Fig.~\ref{fig:train-and-test-phases}.

\subsection{Implicit prior shape model}
The generative model in \cite{AMIRANASHVILI2024}, 
consists of a binary classifier, parameterized by a Multilayer Perceptron (MLP) $f_\theta$, used to classify a voxel as occupied or not given its 3D coordinates $ x \in \mathbb{R}^3 $ and a conditioning vector $ z \in \mathbb{R}^d$:
\begin{equation}
    f_\theta : \mathbb{R}^3 \times \mathbb{R}^d \to [0,1]
\end{equation}
As shown in Fig.~\ref{fig:train-and-test-phases},
the coordinates of all points of the volume under study are given one-by-one to the MLP, along with $z$. The MLP provides the probability  that the voxel associated with the coordinates $x$ is located within the shape, i.e. it predicts its occupancy probability $\hat y = f_\theta(x,z)$.

For a dataset, one latent vector $z_i$ is learned to represent each volume $Y_i$ during training. Such latent conditioning enables to learn a single shared classifier to predict the occupancy grids for the entire population $\{Y_i\}_{i=1}^N$. Learned with an autodecoding scheme, the resultant MLP is a generative model encoding a shape prior over possible values of $z$. In its original version, the shape prior was used to fill incomplete annotations $\hat Y_i$. As we describe next, we deal with full occupancy grids, and an anomaly detection task.

\subsection{Training Phase}
The model $f_\theta$ is trained on full binary 
occupancy grids of a dataset of $N$ ``normal'' muscles $\{Y_i\}_{i=1}^N$. For a training shape $Y_i$, the auto-decoder training scheme is used to compute its latent vector $z_i$. 
In practice, the $\{z_i\}$ are optimized along with the parameters $\theta$. To do so, a Soft Dice loss combined with a cross-entropy loss is used: $L(\hat{Y}_i, Y_i) $. Where $X_i = \{ x_i^j \}_{j=1}^M$ 
are the set of 3D coordinates of each shape, $ Y_i = \{ y_i^j \}_{j=1}^M$ their associated ground truth (GT) occupancies and $\hat{Y}_i = \{ f_\theta(x_i^j, z_i) \}_{j=1}^M$ the predictions. 
The overall minimization problem is formalized as:
\begin{equation}
    \min_{\{z_i\}_{i=1}^N, \theta} \frac{1}{N} \sum_{i=1}^N \left( L(\hat{Y}_i, Y_i) + \lambda \| z_i \|_2^2 \right),
\end{equation}
with $\lambda$ a regularization term.
Effectively, the latent vectors are initialized randomly for each shape $z_i \sim \mathcal{N}(0, 0.1^2)$, and the gradient backpropagation updates both $z_i$ and $\theta$.
\begin{figure*}[hbt!]
  \centering    \includegraphics[height=0.16\paperheight]{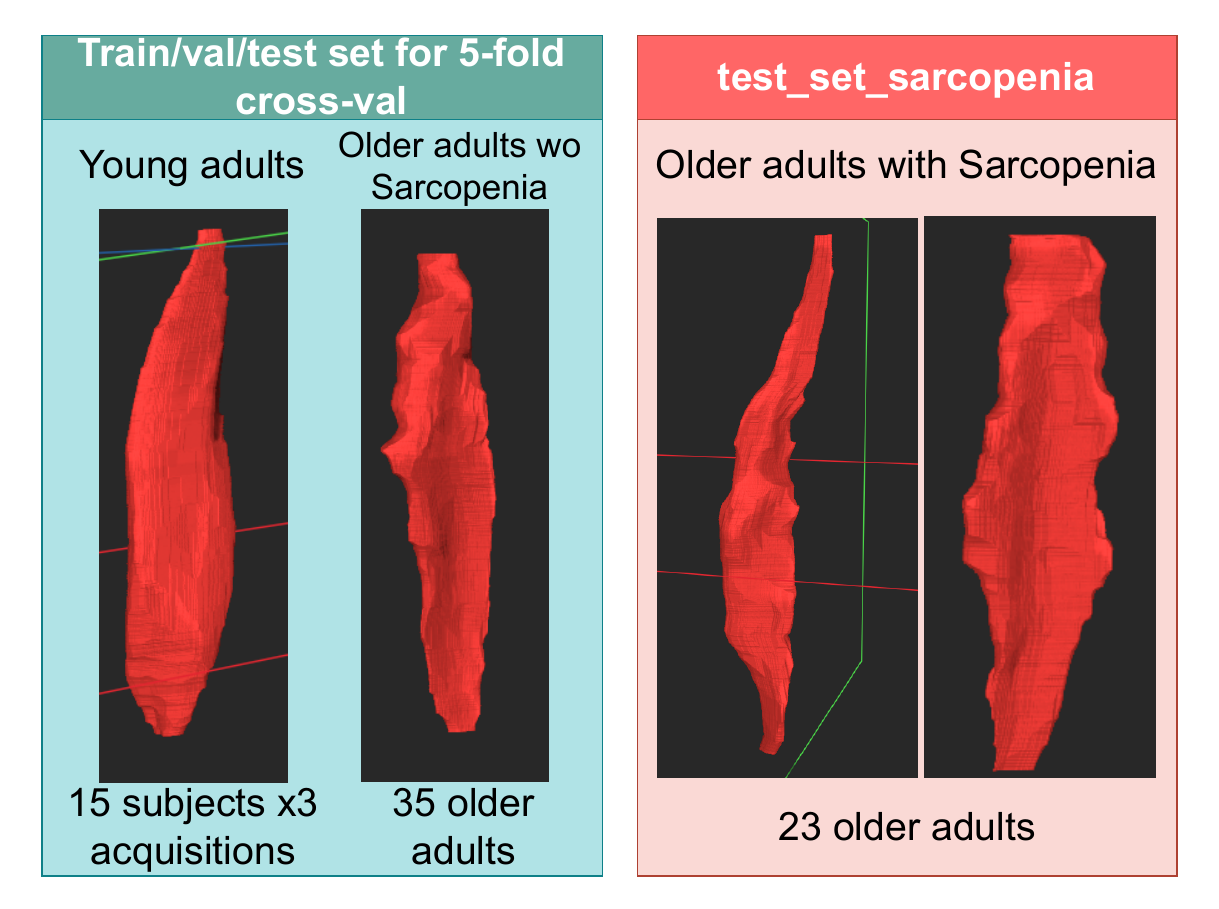}
  \includegraphics[height=0.178\paperheight]{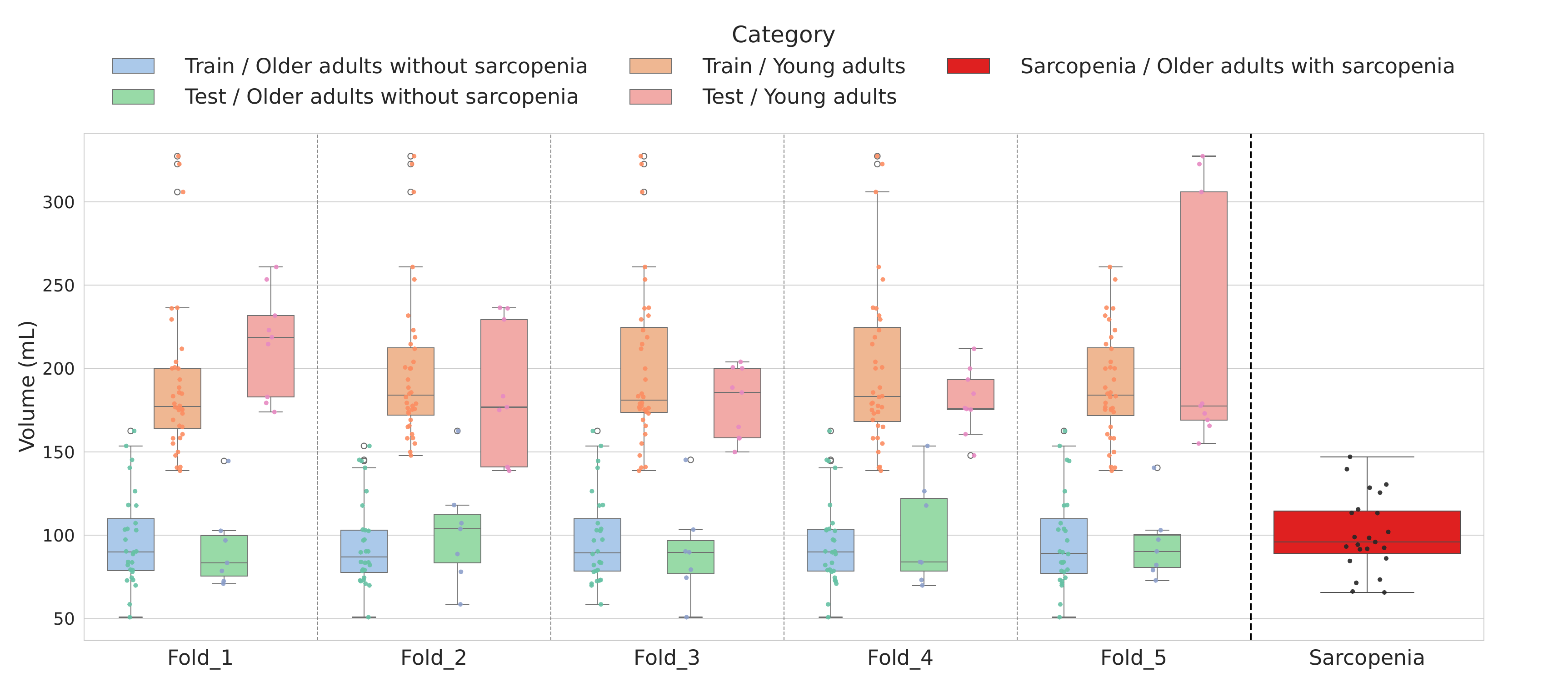} 
  \caption{(left) Example muscles from the dataset. (right) Distribution of the muscle volumes for each fold, considering the train/test split vs. the average volume of RF muscle in patients with sarcopenia.}
  \label{fig:vol_distrib}
\end{figure*}

\subsection{Inference and Anomaly Detection}

During inference (see Fig.~\ref{fig:train-and-test-phases}-bottom),  $f_\theta$ allows to predict new unseen shapes $S$ after training and freezing $\theta$. To this end, the 3D coordinates and full binary GT  occupancy grid $S$ are fed to optimize the unknown latent variable $z_S$. After obtaining the predicted occupancy probabilities $\hat{S}$, 
we minimize:
\begin{equation}
    z_{S} = \arg\min_z \mathcal{L}(\hat{S}, S) + \lambda \| z \|_2^2.
\end{equation}

Since our goal is to discriminate ``normal'' from ``sarcopenic`` muscles, we propose to compute the reconstruction error, here the Dice score $DSC(\hat S, S)$, for a new shape. A threshold on the error is used to detect a ``sarcopenic'' shape.
We also find the projection of all the latent variables ${z_i}$ onto the 2D most discriminant directions to analyse the separation of normal and sarcopenic muscles in the latent space.

\section{Experimental validation}
\label{sec:expe}

\subsection{Experimental settings}
\label{ssec:expe_settings}

\textbf{Datasets.} Two in-house datasets were used for this study. 
The \textit{Young subjects dataset } is composed of 3D Ultrasounds (3DUS)
from 15 healthy participants, 8 males, aged $27 \pm 2$ years (height: $172 \pm 6$ cm, weight: $63 \pm 6 $kg). Each participant was scanned 3 times with a different compression of the probe: standard, minimal compression and using a gel pad; leading to 45 images in total. The segmentations were manually created by an expert clinician. 
The initial study included 4 muscles \cite{HUET2024}, but we focus on the right Rectus Femoris (RF). 
The \textit{DIASEM dataset} includes 58 ``older adults" (age $\geq 75$), hospitalized in rehabilitation or geriatric medicine, with 23 diagnosed with sarcopenia. 3DUS scans were acquired in patients’ rooms using the protocol from \cite{huet2024_journal},  with patients lying on their backs. Three muscles per patient were scanned and manually segmented; as before, we focus on the right RF. Manual segmentations were created by the same expert as for the \textit{Young subjects dataset}. 

The ``normal''  dataset is  composed of all volumes in the \textit{Young subjects dataset } and the non-sarcopenic older subjects of the DIASEM dataset. This normal data are split across the train ($77\%$) and test sets ($23\%$), with an equilibrated repartition of the young ($10\%$) and non-sarcopenic ($13\%$) test muscles. The 23 sarcopenic muscles of \textit{DIASEM} are all part of the test set.  
Experiments follow a 5-fold cross-validation to balance the limited data respecting a subject-wise splitting.

Example RF muscles from each subgroup in Fig.~\ref{fig:vol_distrib}, show that muscles without sarcopenia  can resemble those from sarcopenic older adults. However, sarcopenic muscles appear less ``smooth", with possibly affected boundaries, supporting our hypothesis of characterizing sarcopenia through shape. 
 \\
\textbf{Evaluation Metrics.}
\label{errorCm3}
We measure the volumetric error of the muscles, in $cm^3$ and percentage as:
$\textrm{Vol}_{err_{cm^3}}= |V_{GT}-V_{pred}|$
and 
$\textrm{Vol}_{err_{\%}}= 100\times \frac{|V_{GT}V_{pred}| }{V_{GT}}$,

where $V_{GT}$ and $V_{pred}$ correspond to the ground-truth and predicted volumes of a given muscle respectively. Note that $\textrm{Vol}_{err_{\%}}$ takes into account the size of the muscle while $\textrm{Vol}_{err_{cm^3}}$ does not. 
We also rely on the Dice Score (DSC) to evaluate the reconstruction error. 
\begin{figure*}[bt!]
  {\includegraphics[height=0.16\paperheight]{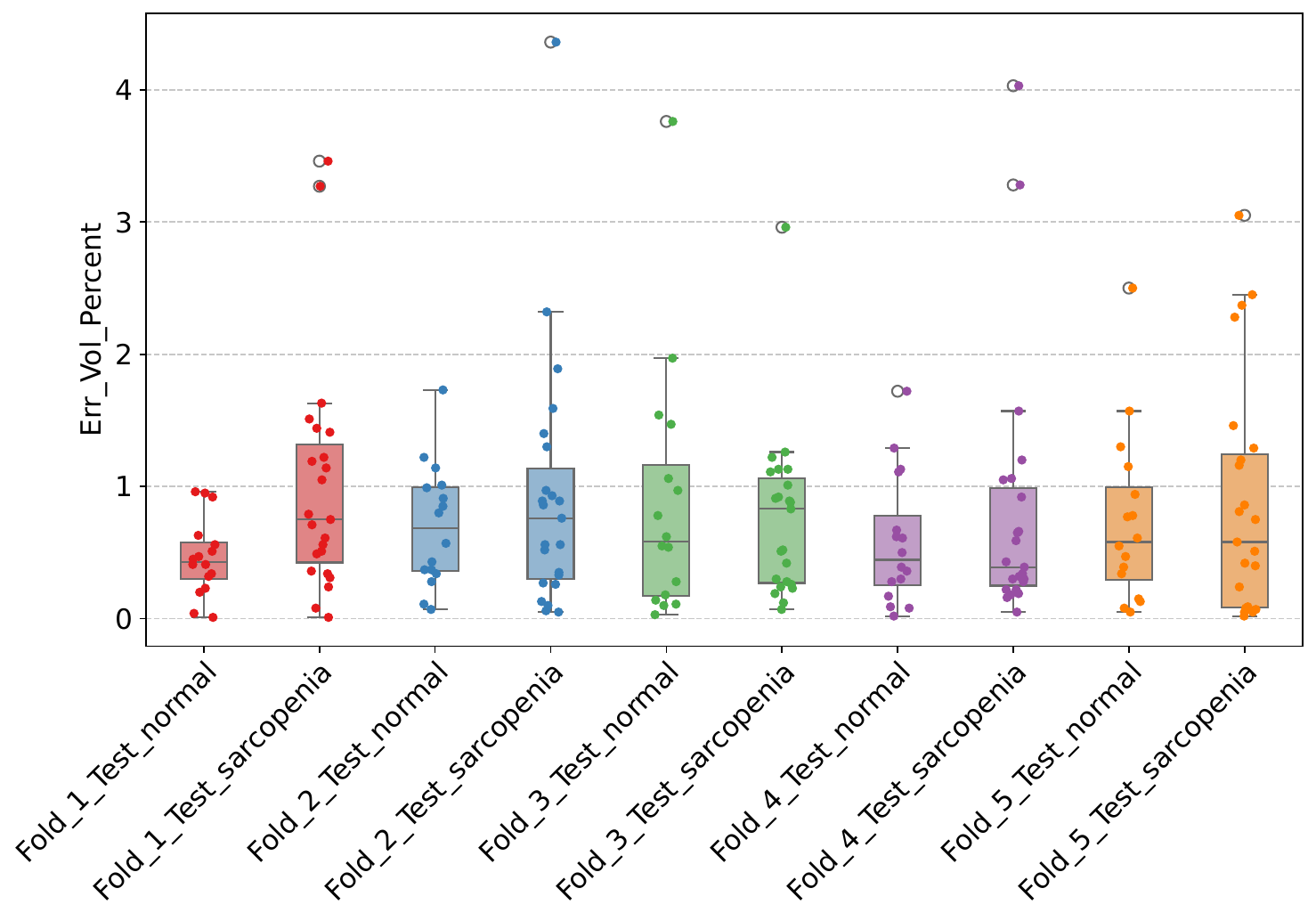}}
  {\includegraphics[height=0.16\paperheight]{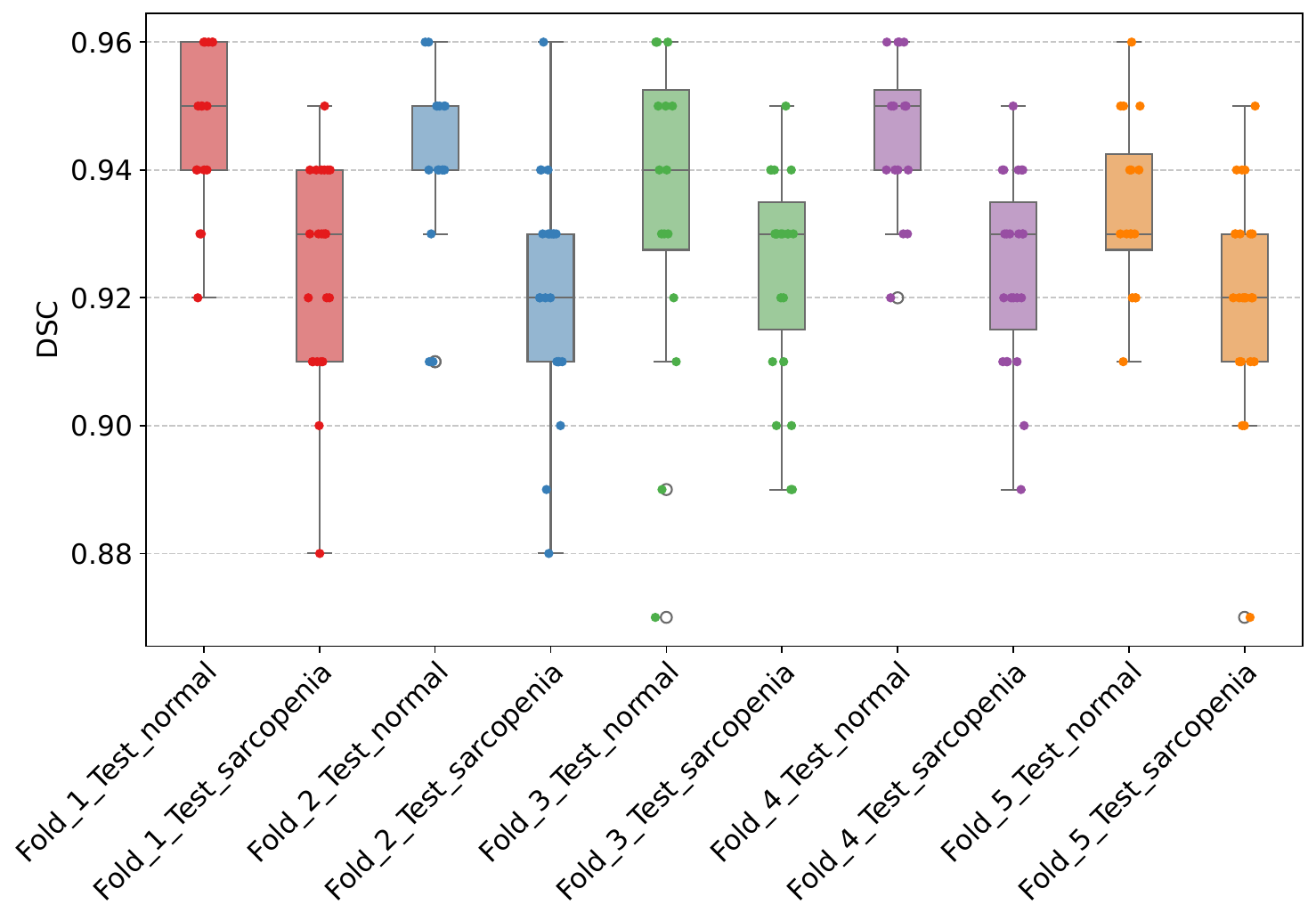}}
   \includegraphics[height=0.175\paperheight]{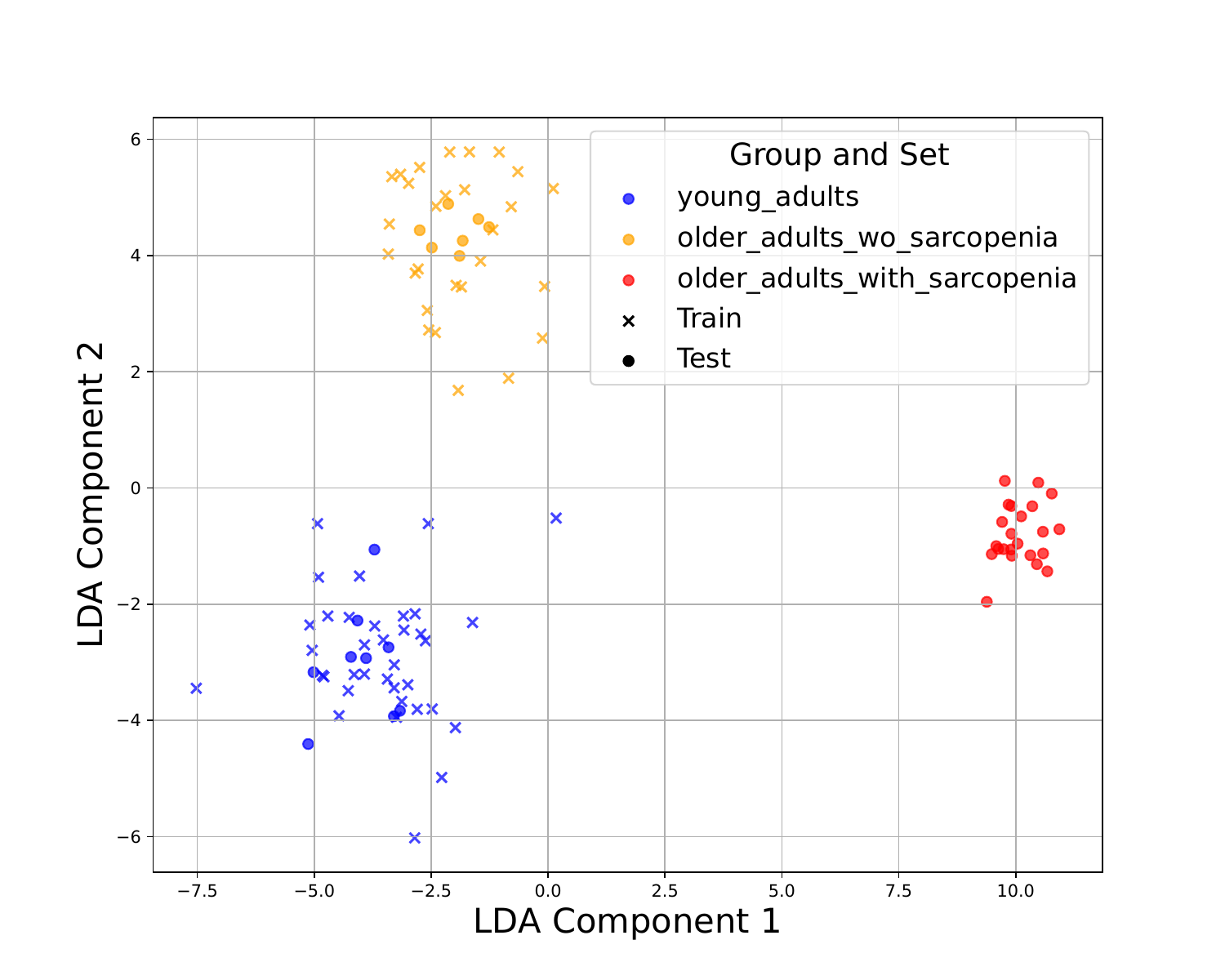} 
   \caption{(a) Volumetric error in \%  and (b) Dice Score by fold for the five test sets separated into  ``Test-healthy''  (young and older adults, non-sarcopenic) and ``Test-sarcopenia`` (sarcopenic). (c) LDA applied on the train and test latents $z$
   of fold 1.}
   \label{fig:quantitativ}
\end{figure*}
\\
\textbf{Implementation details.} \label{impl_details}
Most of the model parameters are kept as in \cite{AMIRANASHVILI2024} including the MLP architecture: the learning rate was set to  $1.0e^{-4}$ and to $1.0e^{-3}$ for the AutoD. The MLP has 8 layers, the output of the 4th layer is concatenated with the 3D coordinates, and the dimension of $z$ is 128.  Contrary to \cite{AMIRANASHVILI2024}, the entire volume was provided during the training and testing phases with a batch size of 1. 
The model was updated using ADAM during 2500 epochs for training and 1500 epochs for inference on a GPU NVIDIA GeForce RTX 3090. Training and inference took $\sim$3h and $\sim$1h, respectively.

\subsection{Quantitative results}
\label{ssec:quantitativ}
\textbf{Muscle volume}
In the first experiment we compare the volume of the muscles as measured in Sec.~\ref{errorCm3} across the different  splits.
The results in Fig. \ref{fig:vol_distrib} show that volume can distinguish subjects from the \textit{Young} and \textit{DIASEM} (older subjects) datasets. In contrast, the volumes of muscles with and without sarcopenia are very close, indicating that a volume threshold alone is insufficient to distinguish between these two populations. Similar observations can be made for the volumes of test muscles reconstructed using $f_\theta$ during inference as reported in Fig.~\ref{fig:quantitativ}-(a). The boxplots show the volumetric error in percent for each fold separated into two subsets: 
``Test\_normal" with non-sarcopenic young and older subjects, and ``Test\_sarcopenia" with sarcopenic  muscles only. While ``normal'' boxplots have a larger variance, the volume of reconstructions is insufficient to characterize sarcopenia.
\\
\textbf{Reconstruction Dice} In contrast to the volume, the proposed reconstruction error based on the Dice error between the original muscles and their reconstruction, as reported in Fig.~\ref{fig:quantitativ}-(b), shows a very relevant separation between the two test groups, suggesting that a threshold of around $0.93$ could detect shape anomalies, i.e. discriminate sarcopenia.

\subsection{Qualitative results and latent space anomalies}
\label{ssec:qualitativ}

 \begin{figure}[h!]
  \centering
  \includegraphics[width=1\linewidth]{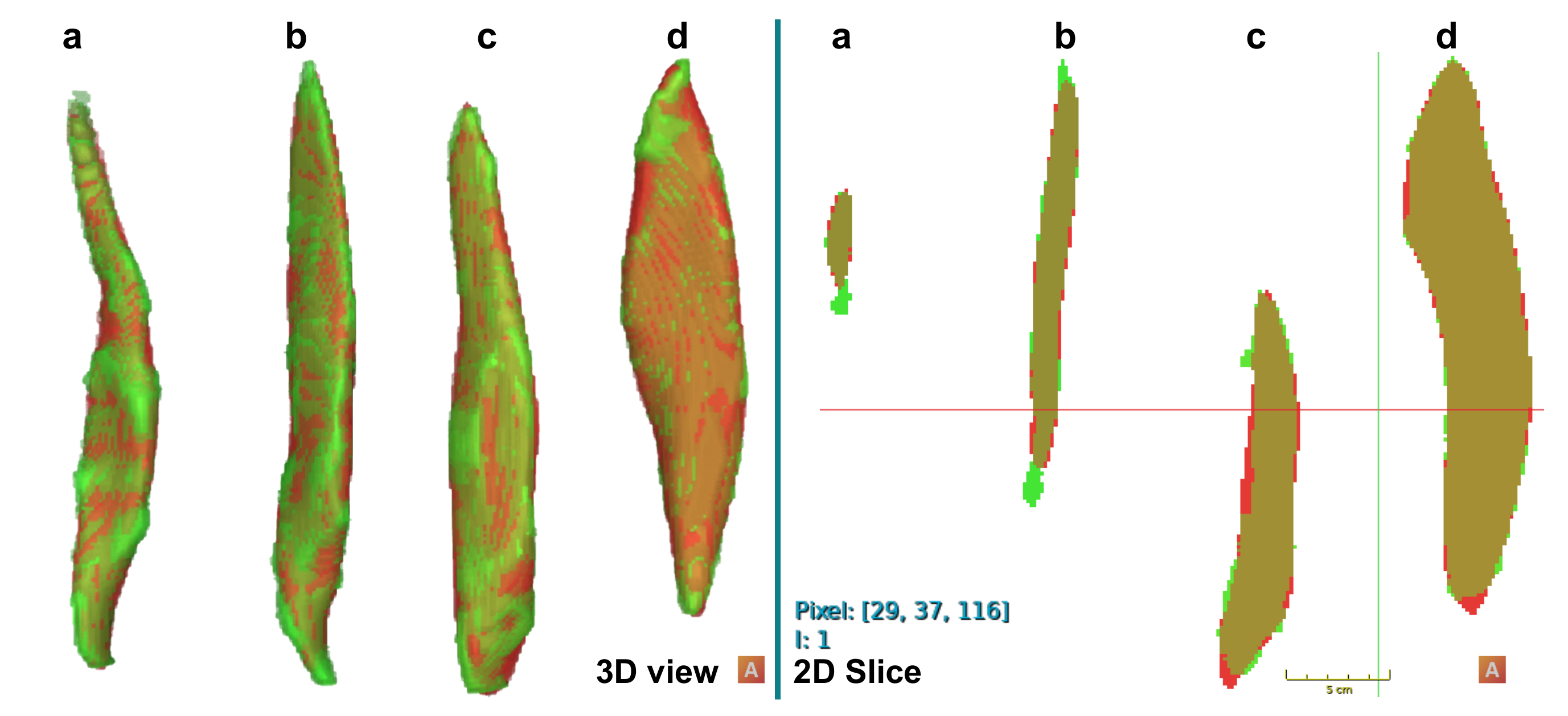} 
  \vspace{-1.5em} 
  \caption{Qualitative results.  Muscles from older adults with (\textbf{a}) and without (\textbf{b}) sarcopenia vs. from young subjects (\textbf{c} and \textbf{d}). Superposition of the GT (in green) and the prediction (in red).}
  \label{fig:qualitativ_imfusion}
\end{figure}

Fig.~\ref{fig:qualitativ_imfusion} shows the occupancy grid prediction of different test subjects in red vs their GT. Overall, the muscles are well reconstructed, even for the older subjects with sarcopenia (\textbf{a}), but some differences are perceptible. These subtle variations are also reflected in the latent space, where differences become much more noticeable. To observe such changes, we apply a Linear Discriminant Analysis (LDA) to the latent spaces of the training and test data of fold 1. We can clearly observe in Fig.~\ref{fig:quantitativ}-(c) the clusters forming according to the populations. It is striking that all normal test data perfectly falls  within the cluster of their training data, while all the sarcopenia muscles fall far away from normal cases.
This suggests that a linear classifier in the latent space can separate the sarcopenic patient population from the rest. If having two normal clusters (yellow and blue) probably captures acquisition variations between the two datasets, the differences between the two DIASEM subpopulations, patients with (red) and without (yellow) sarcopenia, likely represent changes induced by the disease. As a reminder, the two subsets belong to the same DIASEM dataset and follow the same acquisition protocol. Similar results are observed in the remaining folds.

\section{Conclusion}
\label{sec:conclu}

We introduce an implicit neural representation of muscle shapes as well as an unsupervised and implicit shape anomaly detection method based on the representation. To this end, we adapted the method in \cite{AMIRANASHVILI2024}, designed to complete entire volumes from 2D sparse annotations, to a method for anomaly detection. Training on a dataset of non-sarcopenic muscles, we learned a shape prior that captures variations within the training population. Consequently, out-of-distribution muscles exhibit higher reconstruction errors, enabling the detection of sarcopenic muscles. Experimental results on 103 segmented muscles indicate lower Dice reconstruction errors for sarcopenic muscles. Additionally, an LDA analysis of the latent shape representations effectively clusters non-sarcopenic populations and distinctly separates sarcopenic muscles during inference. In future work, we plan to apply the proposed method to other muscles and datasets. Additionally, incorporating an automatic segmentation could address the challenges associated with manual annotations.
Lastly, a more thorough analysis of the learned latent representations could improve interpretability of the captured differences.

\section{Compliance with ethical standarts}
\label{sec:ethic}
This protocol was approved by the local ethics committee of the Nantes University Hospital "Groupe Nantais d'éthique dans le Domaine de la Santé" : AVIS 24-6-01-191. 


\bibliographystyle{IEEEbib}
\bibliography{biblio-macro,refs}

\end{document}